\documentclass[11pt]{article}
\usepackage{acl}
\usepackage{times}
\usepackage{latexsym}
\usepackage{caption}
\usepackage[T1]{fontenc}
\usepackage[utf8]{inputenc}
\usepackage{microtype}
\usepackage{inconsolata}
\usepackage{graphicx}
\usepackage{hyperref}       
\usepackage{url} 
\usepackage{booktabs}
\usepackage{amsfonts} 
\usepackage{nicefrac}           
\usepackage{xcolor}         
\usepackage{natbib}
\usepackage{sidecap}
\usepackage{float}
\usepackage{afterpage}

\title{\ourmethod: Efficient Watermarking for Diffusion Language Models}

\usepackage{algorithm}
\usepackage[noend]{algpseudocode}
\usepackage{amsmath}

\newcommand{\ourmethod}{LR-DWM}

\author{
Ofek Raban\thanks{Corresponding author: \texttt{ofekraban@gmail.com}} \\
Bar-Ilan University\
\And
Ethan Fetaya \\
Bar-Ilan University
\And
Gal Chechik \\
Bar-Ilan University\\
NVIDIA
}
\begin{document}
\maketitle
\begin{abstract}
Watermarking (WM) is a critical mechanism for detecting and attributing AI-generated content. Current WM methods for Large Language Models (LLMs) are predominantly tailored for autoregressive (AR) models: They rely on tokens being generated sequentially, and embed stable signals within the generated sequence based on the previously sampled text. Diffusion Language Models (DLMs) generate text via non-sequential iterative denoising, which requires significant modification to use WM methods designed for AR models.
Recent work proposed to watermark DLMs by inverting the process when needed, but suffers significant computational or memory overhead. 

We introduce \textit{Left-Right Diffusion Watermarking} (\emph{\ourmethod}), a scheme that biases the generated token based on both left and right neighbors, when they are available.
\ourmethod{} incurs minimal runtime and memory overhead, remaining close to the non-watermarked baseline DLM while enabling reliable statistical detection under standard evaluation settings. Our results demonstrate that DLMs can be watermarked efficiently, achieving high detectability with negligible computational and memory overhead.

\end{abstract}

\section{Introduction}

\begin{figure}[t]
    \centering
    \includegraphics[width=0.98\linewidth]{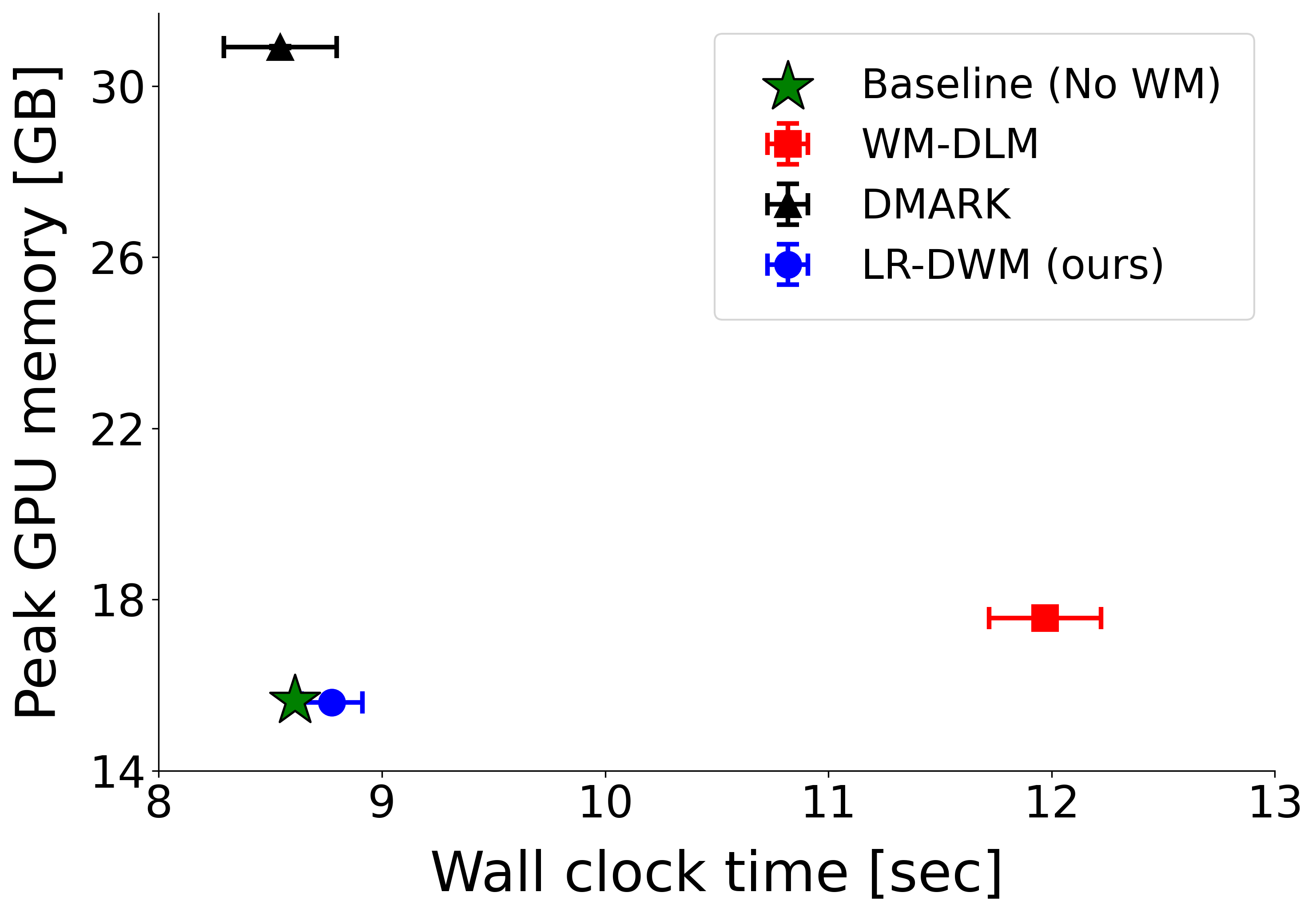}
    \caption{\textbf{Computational Efficiency.}
Peak GPU memory versus wall-clock generation time on a single NVIDIA H100.
The non-watermarked baseline (green star) is shown for reference.
}

    \label{fig:efficiency_plot}
\end{figure}

Watermarking has become a common technique for detection and attribution of text generated by Large Language Models (LLMs). Most existing watermarking schemes are designed for autoregressive (AR) models and critically rely on a simple but powerful property: tokens are generated in a fixed, deterministic left-to-right order. This known ordering allows the watermarking algorithm to use the previously sampled tokens to bias the generation of the next token, using a hash function to favor tokens from a “green” set. A detector can then count the frequency of “green” tokens and flag the text as LLM-generated if they are overrepresented in the text. 

Diffusion Language Models (DLMs), receive increasing attention as a low-latency alternative to AR models, but they break the sequential assumption at its core. Instead of exposing tokens in a fixed left-to-right order, they generate text by iteratively denoising an entire sequence in a non-sequential order. At each step, different positions may be updated, and the schedule by which tokens are determined is not known in advance. As a result, one cannot bias a generated token based on tokens on its left, because they may still be undecided when we sample the current token. In this setting, the standard AR watermarking toolkit is invalid without substantial modification. Since DLMs offer unique advantages in parallel decoding, generation speed, and fine-grained contextual control, there is a need for reliable watermarking in this paradigm.

Very recently, \cite{gloaguen2025wmdlm, wu2025dmark} adapted AR watermarking to DLMs by trying to invert the hashing function: Be ``green” for the previous token, and also make the next token ``green”. This, however, incurs significant computational or memory overhead. In contrast, we propose a WM scheme that can be implemented with minimal runtime and memory overhead relative to the evaluated baseline.

Our key observation is that since generation order in diffusion models is no longer sequential, watermark constraints need not be restricted to a single causal direction. Instead, a token can be independently influenced by both its left and right neighbors whenever they are available. Based on this observation, we introduce \emph{\ourmethod}: During generation, \emph{\ourmethod} biases the logits in an additive way based on  available neighboring tokens from each direction. This design naturally favors tokens that satisfy both left and right constraints, while remaining compatible with the unknown order of diffusion decoding.

The main contributions of this paper are: 

\noindent \textbf{(1)} A novel two-sided watermarking method for diffusion language models. It embeds watermark signals by independently leveraging both left and right local context in an order-agnostic manner.

\noindent \textbf{(2)} We show that \emph{\ourmethod} enables \emph{efficient} watermarking for DLMs, incurring negligible runtime and memory overhead, while maintaining competitive detectability and text quality.

\section{Technical Background}

\paragraph{\textbf{Watermarking for Language Models}}
Watermarking for language models embeds a statistical signal by biasing token sampling during decoding using a secret key, typically by hashing previously generated tokens to partition the vocabulary into preferred (green) and non-preferred (red) sets. Then, boosting the green tokens yields a bias that can later be detected \cite{kirchenbauer2023watermark,hu2023variance,zhao2023finite}.
A key objective is to balance detectability against quality degradation \cite{tu2024waterbench,pan2024robust}.

\paragraph{\textbf{Diffusion Language Models}}
Diffusion Language Models (DLMs) generate text by iteratively denoising a corrupted token sequence \cite{ho2020ddpm,song2020score,austin2021d3pm,lou2023diffusion,sahoo2024mdlm}.
Unlike AR models, DLM decoding updates tokens in a non-monotonic order, enabling parallel refinement and faster generation, and supporting both stochastic and deterministic decoding \cite{ye2025dream,kim2025entropy,nie2025llada}.
Since generation does not follow a fixed left-to-right factorization, watermarking methods developed for AR decoding are not directly applicable to DLMs.
\section{Related Work}

\paragraph{\textbf{Watermarking Diffusion Language Models}}
Recent work begun extending watermarking to diffusion language models (DLMs).

WM-DLM \cite{gloaguen2025wmdlm} embeds watermarks by marginalizing, in expectation, over missing neighboring tokens during diffusion steps; this incurs substantial per-step computation due to repeated expectation and requires non-zero sampling temperature, as deterministic decoding ($T=0$) eliminates sampling randomness and prevents watermark application.\\

DMARK \cite{wu2025dmark} proposes an order-agnostic watermarking strategy that adapts AR-style watermark constraints to non-causal decoding via an inverse-style mechanism. When the token to the right of the currently decoded token is known, we bias the tokens that would make the right token green. In order to avoid computational overhead during generation, the method relies on $|\mathcal{V}|^2$ cached lookup tables, trading memory usage for runtime efficiency.

\section{Problem Setup}
We consider the problem of watermarking text generated by a Diffusion Language Model (DLM). The input to the system is a user-provided prompt $c$, and the output is a generated text sequence $y = (y_1, \ldots, y_T)$ from a vocabulary $\mathcal{V}$ sampled in an unknown order. 

\subsection{Watermarking Framework}
To address the non-sequential nature of diffusion, we view watermarking as a set of local constraints defined over the final token sequence.
At each denoising step $t$, the model predicts token distributions for positions that have not yet been finalized, and we modify these distributions via lightweight logit biasing. Specifically, let $l_v \in \mathbb{R}^{|\mathcal{V}|}$ denote the logit for token $v$; we add a bias $\delta$ whenever a watermark constraint induced by neighboring tokens is available.
Importantly, these constraints are defined solely from the previously determined token. As a result, detection is schedule-agnostic and does not try to invert the hashing process.

\subsection{Left-Right Diffusion Watermarking}
\label{subsec:\ourmethod}
\begin{figure}[t]
    \centering
    \includegraphics[width=0.95\linewidth]{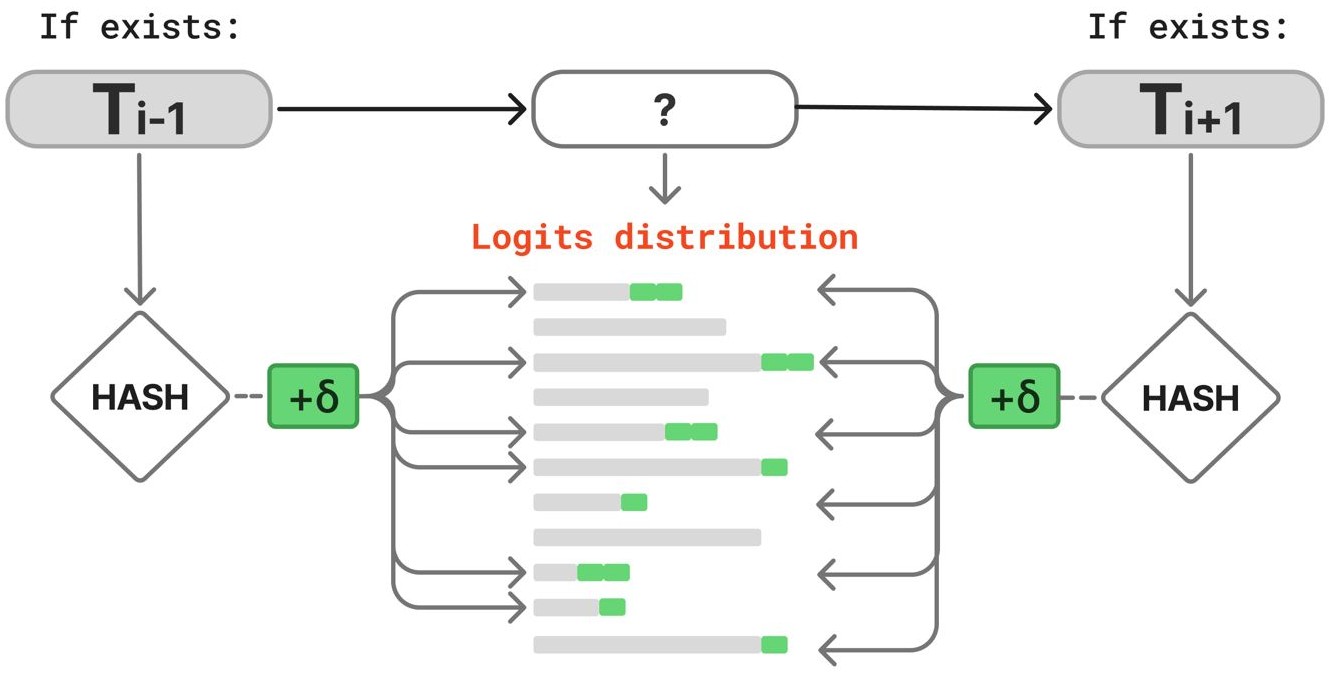} 
    \caption{\textbf{Schematic of \ourmethod{}.}
For a target token $y_i$, the model’s logits distribution is modified using bidirectional local context.
If available, the left neighbor $y_{i-1}$ and the right neighbor $y_{i+1}$ independently induce green-token constraints via hash functions with distinct keys $k_L$ and $k_R$.
Each constraint applies an additive bias $\delta$ to its corresponding green set.
Boundary tokens fall back to a single-sided constraint when one neighbor is missing.}
    \label{fig:schematic}
\end{figure}

We introduce \emph{\ourmethod}, an algorithm that induces \emph{efficient} watermarking by anchoring signals to the known local neighbors.

\paragraph{Bidirectional Context Hashing.}
For a token at position $i$, we treat the left neighbor ($y_{i-1}$) and right neighbor ($y_{i+1}$) as independent sources of watermark signal. We hash twice, with secret keys $k_L$ and $k_R$, to compute the set of green tokens:
\begin{equation}
    G_L^{(i)} = \mathcal{H}(y_{i-1}, k_L), \quad G_R^{(i)} = \mathcal{H}(y_{i+1}, k_R)
\end{equation}
where $\mathcal{H}$ uses a cryptographic hash function \cite{kirchenbauer2023watermark}. When a neighbor is unavailable, the green set is empty.

\paragraph{Injection Strategy.}
 Since the constraints are applied independently, the total logit modification is the sum of individual biases:
\begin{equation}
    l'_v = l_v + \delta \cdot \mathbb{I}(v \in G_L^{(i)}) + \delta \cdot \mathbb{I}(v \in G_R^{(i)})
\end{equation}
This additive mechanism naturally favors tokens that satisfy both constraints (when applicable), thereby maximizing the probability of generating tokens that yield a high positive detection score.

\subsection{Detection Statistic}
As each token has left and right watermarking signals, we unify the signals into a single ternary score $s_i$. Let $m_L = \mathbb{I}(y_i \in G_L^{(i)})$ and $m_R =\mathbb{I}(y_i \in G_R^{(i)})$ denote whether token $y_i$ matches the green list induced by the left or right neighbor, respectively. We define the score for token $y_i$ as:
\begin{equation}
    s_i = m_L + m_R - 1 \; ,
\end{equation}
yielding $s_i \in \{+1, 0, -1\}$.
A score of $+1$ indicates token matches both neighbors (positive signal), $0$ corresponds to a single-sided match (neutral), and $-1$ indicates no match (negative signal).
Under the null hypothesis $H_0$  of the text being written by a person, token membership in green lists is random. Since the hash function partitions the vocabulary uniformly into two balanced sets, each match indicator is a Bernoulli random variable, yielding an expected score of zero: $\mathbb{E}[s_i] = 0$.
It follows that $\mathrm{Var}(s_i)=1/2$ under the random-hash null. While adjacent scores may be weakly dependent
due to shared neighboring tokens, this dependence is local and handled via empirical calibration of $\sigma^2$ on human-written text.\\

The final detection statistic is the standardized sum over a sequence of length $T$:
\[
Z = \frac{1}{\sigma \sqrt{T}} \sum_{i=1}^T s_i,
\]
where $\sigma^2$ is estimated from human-written text. Under $H_0$, the resulting statistic is empirically well-approximated by a normal distribution, enabling standard Z-score–based thresholding with calibrated false positive rates.

To validate the statistical assumptions underlying our detection algorithm, we empirically evaluate the null hypothesis $H_0$ using 10{,}000 human-written texts of length 400 tokens sampled from the C4 corpus~\citep{raffel2020exploring}. We set a threshold on the Z-score that empirically  corresponds to 1\% FPR and ensure that indeed we get 1\% error rate on this human-written corpus.

\section{Experiments}
\label{sec:experiments}

\subsection{Experimental Setup}

\paragraph{Model and Generation Setup.}
We evaluate our method on two state-of-the-art diffusion language models (DLMs):
\textbf{LLaDA-8B-Instruct}~\cite{nie2025llada} and \textbf{DREAM-7B-Instruct}~\cite{ye2025dream}.
For each model, we adhere to the authors' recommended configurations, representing
two distinct decoding paradigms: (1) \textbf{LLaDA} employs deterministic decoding
with greedy refinement and a block length of 25; (2) \textbf{DREAM} uses its
standard stochastic decoding setup. In both cases, we generate
300-token sequences over 300 diffusion steps. Results on \textbf{DREAM} are in the supplementary material.

\paragraph{Datasets.}
We used the 600 prompts from  \textbf{WaterBench}~\cite{tu2024waterbench}.
To ensure consistency with prior DLM watermarking evaluations and to mitigate
diffusion-specific degeneration effects (e.g., repetitive loops), we adopt the
post-generation filtering protocol of \citet{gloaguen2025wmdlm}.

\paragraph{Metrics.}
We evaluate watermarking performance along three axes:
\textbf{(1) Detectability,} measured as true positive rate (TPR) at a 1\% false positive
rate (FPR), following prior watermarking work, with human-written text treated as negatives;
\textbf{(2) Text Quality,} measured via perplexity (PPL) using \textbf{Qwen2.5-32B}~\cite{qwen2024technical}
as an external oracle

\textbf{(3) Efficiency,} measured by wall-clock generation time and peak GPU memory usage
on a single \textbf{NVIDIA H100} GPU.

\paragraph{Baselines.}
We compare \textbf{\ourmethod} against the unmodified \emph{vanilla} models and two recent
diffusion watermarking methods: \textbf{WM-DLM}~\cite{gloaguen2025wmdlm}, a stochastic
expectation-based approach, and \textbf{DMARK}~\cite{wu2025dmark}, a deterministic
kernel-based method. We note that WM-DLM performance strongly depends on stochastic decoding (with high temperature), however for both LLaDa and DREAM DLMs, this means the base model it watermarks has considerably worse performence. 

\subsection{Main Results: Computational Efficiency }
\label{subsec:efficiency}
Figure~\ref{fig:efficiency_plot} reports absolute wall-clock generation time and peak GPU memory usage for each method under identical experimental settings.
The non-watermarked baseline is included for reference and exhibits identical efficiency under deterministic and stochastic decoding.\\

\ourmethod{} introduces minimal additional cost, remaining close to the non-watermarked deterministic baseline in both wall-clock time and memory usage.
In contrast, DMARK incurs a substantially larger memory footprint due to caching the entire hash table in advance, nearly doubling peak GPU memory consumption.
WM-DLM exhibits higher runtime overhead, reflecting the cost of its expectation-based
scoring under stochastic decoding.

Overall, these results indicate that \ourmethod{} achieves watermarking with
negligible computational overhead to its base generation configuration, which is the primary focus of this work.

\paragraph{Detectability vs Naturalness tradeoff}
\begin{table}[t]
\centering
\caption{Effectiveness validation: estimated perplexity (PPL $\pm$ SEM) at three fixed detection rates. }
\label{tab:ppl_detection}
\resizebox{\columnwidth}{!}{%
\begin{tabular}{lccc}
\toprule
\textbf{Detection rates} & \textbf{90\%} & \textbf{99\%} & \textbf{99.5\%} \\
\midrule
WM-DLM 
& 5.07 $\pm$ 1.40 
& 6.14 $\pm$ 1.84 
& 6.33 $\pm$ 1.90 \\

DMARK 
& 2.82 $\pm$ 0.51 
& 3.28 $\pm$ 0.61 
& 3.34 $\pm$ 0.63 \\

\midrule
{LR-DWM (OURS)} 
& 2.80 $\pm$ 0.46 
& 3.32 $\pm$ 0.65 
&  3.37 $\pm$ 0.66 \\
\bottomrule
\end{tabular}}
\end{table}

We next verify that the efficiency gains of \ourmethod{} do not come at the expense of watermark effectiveness.
Table~\ref{tab:ppl_detection} reports perplexity at fixed detection operating points on LLaDA,
showing that \ourmethod{} maintains competitive detectability and text quality relative to existing
diffusion watermarking methods, see full curve in the Appendix.
On DREAM-7B, we observe a similar quality-detectability trade-off, with comparable perplexity
at matched detection rates (Appendix Figure~\ref{fig:tradeoff_ppl_detection}).

\subsection{Watermark Robustness}
We evaluate robustness using the \textsc{MarkLLM} benchmark~\citep{pan2024markllm} under standard non-adaptive text perturbations.
At a fixed operating point achieving $100\%$ detection on clean text, LR-DWM retains high detection rates under most moderate lexical attacks, including $98.8\%$ detection under $10\%$ word deletion and $99.4\%$ under $10\%$ context-aware word substitution. The substantial drop under paraphrasing reflects the disruption of the local bidirectional lexical context exploited by LR-DWM, a known limitation shared by watermarking methods based on local token correlations.\\

Detailed robustness results across attack types and intensities are reported in Appendix~A (Table~\ref{tab:robustness_appendix}).

\section{Conclusion}
We introduced \ourmethod{}, a lightweight watermarking scheme designed for diffusion LMs that does not rely on tokens being generated in sequential order. In contrast with existing methods, \ourmethod{} is both  fast and memory efficient:  
Across two diffusion models, \ourmethod{} demonstrates minimal encoding and inference overhead relative to the evaluated baselines, while maintaining detectability and naturalness comparable to existing diffusion watermarking methods under standard non-adaptive evaluation settings.

\clearpage
\subsection*{Limitations}
Our watermarking method has several limitation, that are shared with other current watermarking methods. While the decrease in text quality is small, but this decrease in performance can be non-negligible if one requires very high detection rates. Watermarking also requires a minimal text length for the signal to be statistically significant. Moreover,  robustness evaluation focuses on standard text perturbations, and not adversarial attacks.

\bibliography{custom}
\clearpage

\section{Appendix}
\raggedbottom
\subsection{Algorithm Pseudocode}
This section provides additional algorithmic details for \ourmethod{},
complementing the high-level description in the main paper.

\begin{algorithm}[H]
\small 
\caption{\textbf{\ourmethod} Diffusion Decoding with Two-Sided Watermarking (Simplified)}
\label{alg:\ourmethod-simple}
\begin{algorithmic}[1]
\Require Prompt $c$, Model $\mathcal{M}$, Keys $k_L, k_R$, Bias $\delta$
\Ensure Generated sequence $y$

\State Initialize $y^{(S)}$ as a fully masked or corrupted token sequence
\For{$s = S, \dots, 1$}
    \State $\mathcal{I}_s \gets$ positions selected for update by the diffusion schedule at step $s$
    \For{each position $i \in \mathcal{I}_s$}
        \State $\ell \gets \mathcal{M}(y, i)$ \Comment{Base logits}
        
        \If{$y_{i-1}$ is revealed}
            \State $M_L \gets \textsc{GreenMask}(y_{i-1}, k_L)$
            \State $\ell \gets \ell + \delta \cdot M_L$
        \EndIf  
        
        \If{$y_{i+1}$ is revealed}
            \State $M_R \gets \textsc{GreenMask}(y_{i+1}, k_R)$
            \State $\ell \gets \ell + \delta \cdot M_R$
        \EndIf
        
        \State $y_i \sim \textsc{Decode}(\ell)$
    \EndFor
\EndFor
\State \Return $y$
\end{algorithmic}
\end{algorithm}

\vspace{-5pt} 
\noindent
\begin{minipage}{\linewidth}
    \subsection{Detailed Results}
    \subsubsection{Trade-off Curves comparison on LLaDA}
    \begin{center}
        \includegraphics[width=0.85\linewidth]{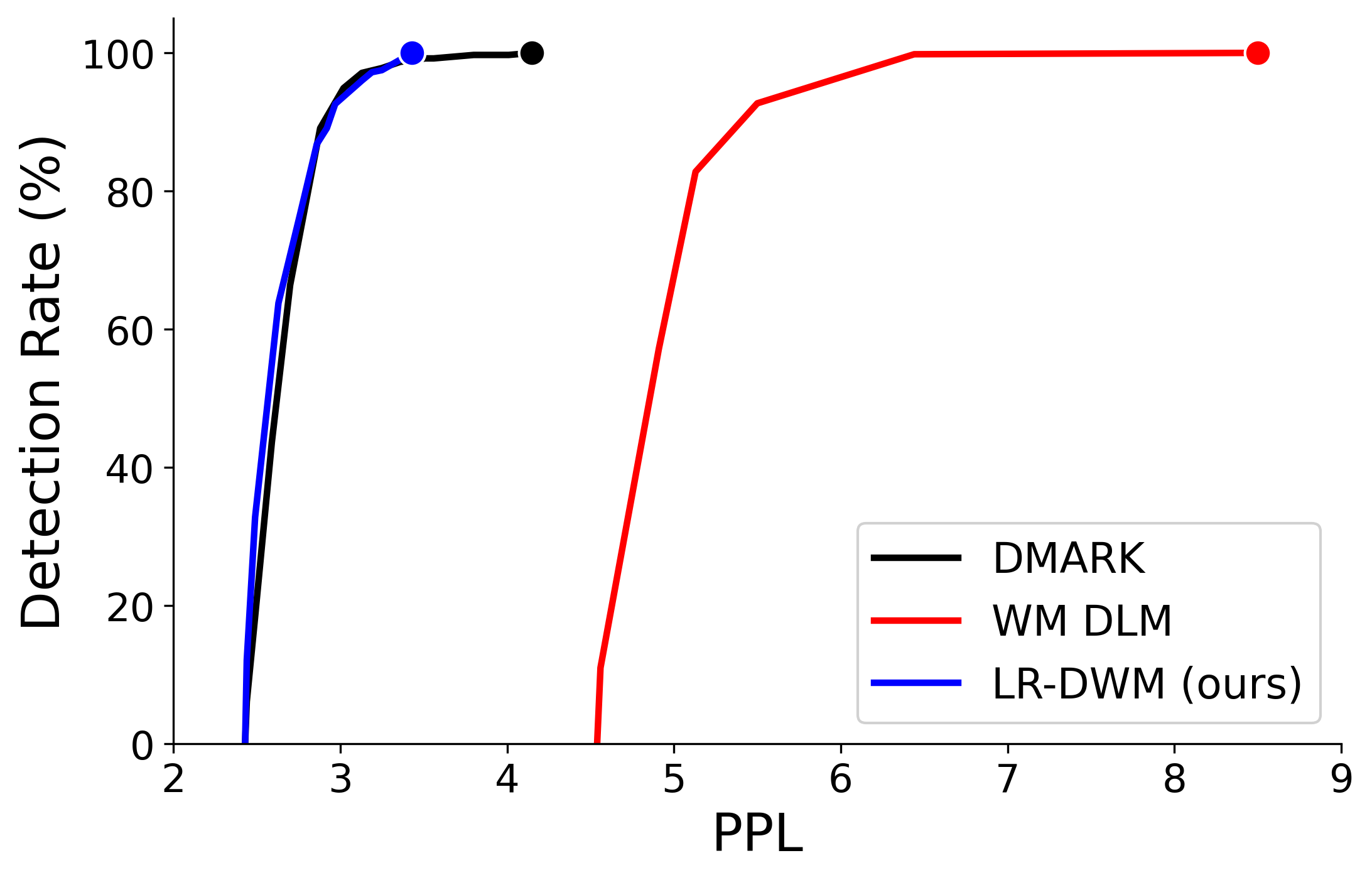}
        \captionof{figure}{\textbf{Quality-Detectability Trade-off.} Detection rate at a FPR of 1\% versus perplexity (PPL) across a range of bias strengths $\delta$. {\ourmethod} exhibits a competitive trade-off relative to DMARK and WM-DLM.}
        \label{fig:tradeoff_ppl_detection}
    \end{center}
\end{minipage}

\subsubsection{Quality-Detectability Trade-off on DREAM}

\begin{figure}[H]
    \centering
    \includegraphics[width=0.98\linewidth]{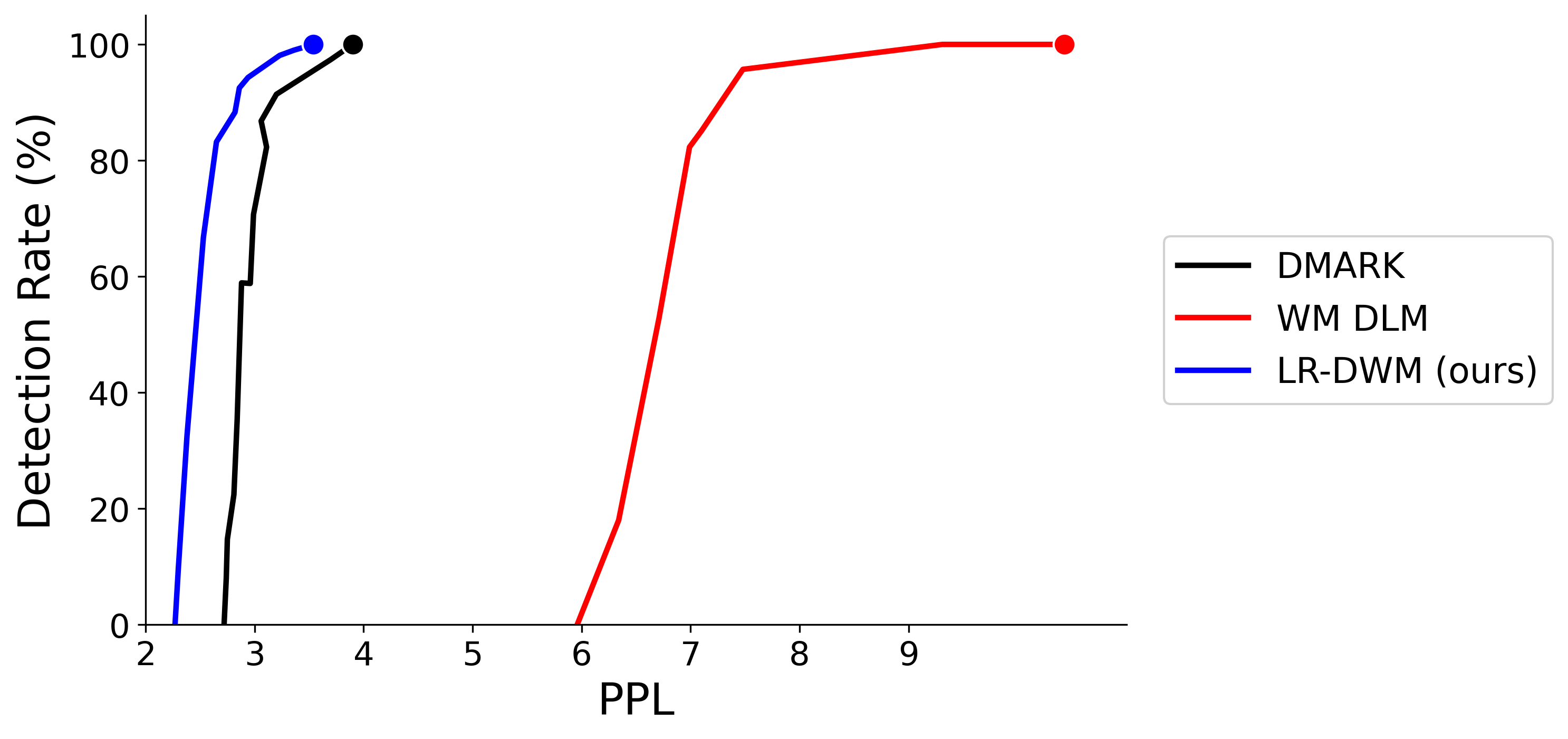}
    \caption{
    \textbf{Quality-Detectability Trade-off on DREAM.}
    Detection rate versus perplexity (PPL) on the DREAM model for LR-DWM
and prior watermarking methods, evaluated under stochastic decoding with low temperature.
Each point corresponds to a different watermark strength~$\delta$.
LR-DWM exhibits a sharp detectability transition at lower PPL values,
closely matching the trend observed on LLaDA and indicating consistent
generalization across diffusion-based language models.
    }
    \label{fig:dream_tradeoff}
\end{figure}
\paragraph{Generation Setup for DREAM.}
Results for LR-DWM and DMARK shown in Figure~\ref{fig:dream_tradeoff}
were obtained using the decoding configuration recommended by the DREAM authors.
We generate sequences of length 300 using 300 diffusion steps,
with entropy-based stochastic decoding and a deliberately low temperature setting.
Specifically, we use:
\texttt{GEN\_LENGTH}=300,
\texttt{STEPS}=300,
\texttt{TEMPERATURE}=0.2,
\texttt{ALG}=\texttt{entropy},
\texttt{ALG\_TEMP}=0.0,
\texttt{EPS}=1e{-}3,
\texttt{TOP\_P}=0.95,
and no top-$k$ truncation.

Due to the combination of low-temperature decoding and diffusion-based sampling,
generation quality is more sensitive to standard quality and length filters,
and a substantial fraction of generated texts did not pass these filters.
All detection results are therefore computed over the surviving subset.
We apply the same generation and filtering procedure consistently across
LR-DWM and DMARK on DREAM.

For WM-DLM, the decoding configuration used above did not produce
stable generations.
We therefore report WM-DLM results using the decoding setup recommended
in the original WM-DLM work, while keeping all evaluation metrics and
filtering criteria identical.

Finally, we observe a larger gap between LR-DWM and DMARK on DREAM than on LLaDA.
We attribute this difference primarily to decoding dynamics and filtering
effects under low-temperature diffusion sampling, rather than to changes
in the underlying watermark signal.

\begin{table*}[t]
\centering
\caption{\textbf{Robustness of LR-DWM under non-adaptive attacks.}
Detection performance under standard text perturbations at a fixed false positive rate (FPR) of 1\%.
The baseline corresponds to clean watermarked text generated with $\delta=3.25$, achieving 100\% detection and an average Z-score of 6.653.
Reported drops are relative to this baseline.
Back-translation uses local OPUS-MT models (EN$\rightarrow$ZH$\rightarrow$EN).}

\label{tab:robustness_appendix}
\begin{tabular}{llcccc}
\toprule
\textbf{Attack Type} & \textbf{Param.} & \textbf{Detection (\%)} & \textbf{Drop (\%)} & \textbf{Avg. Z} & \textbf{Z Drop} \\
\midrule
{Word deletion}
& 10\% & 98.83 & 1.17 & 5.45 & 1.20 \\
& 20\% & 96.78 & 3.22 & 4.74 & 1.91 \\
& 30\% & 90.94 & 9.06 & 3.91 & 2.74 \\
& 40\% & 74.85 & 25.15 & 3.13 & 3.52 \\
& 50\% & 59.06 & 40.94 & 2.55 & 4.10 \\
\midrule
{Word substitution}
& 10\% & 98.25 & 1.75 & 5.29 & 1.36 \\
& 20\% & 95.32 & 4.68 & 4.49 & 2.16 \\
& 30\% & 86.84 & 13.16 & 3.73 & 2.92 \\
& 40\% & 70.47 & 29.53 & 3.06 & 3.59 \\
& 50\% & 51.46 & 48.54 & 2.34 & 4.31 \\
\midrule
{Context-aware substitution (BERT)}
& 10\% & 99.42 & 0.58 & 5.83 & 0.82 \\
& 20\% & 98.83 & 1.17 & 5.48 & 1.18 \\
& 30\% & 98.54 & 1.46 & 5.31 & 1.34 \\
\midrule
Back-translation(EN$\rightarrow$ZH$\rightarrow$EN)
& - & 55.56 & 44.44 & 2.48 & 4.17 \\
\midrule
Paraphrasing (Qwen-2B) 
& - & 15.79 & 84.21 & 1.02 & 5.63 \\
\bottomrule
\end{tabular}
\end{table*}

\subsection{Robustness to Non-adaptive Attacks}
 
We evaluate robustness under standard \emph{non-adaptive} text perturbations using the MARKLLM benchmark.
Table~\ref{tab:robustness_appendix} reports detection performance at a fixed operating point calibrated to a 1\% false positive rate (FPR) on clean text.

\emph{Paraphrasing is the most damaging attack}, substantially reducing detectability by altering local lexical structure.
In contrast, \emph{light attacks} such as word deletion and random substitution preserve high detection rates, indicating a distributed watermark signal.

Robustness under \emph{context-aware substitution} is partly explained by a benchmark bottleneck, as WordNet-based substitutions affect only tokens with available synonyms.

\end{document}